\pdfoutput=1

\documentclass[11pt]{article}

\usepackage{naacl2021}

\usepackage{times}
\usepackage{latexsym}

\usepackage[T1]{fontenc}

\usepackage[utf8]{inputenc}

\usepackage{microtype}

\usepackage{xspace}
\usepackage{amsmath}
\usepackage{amssymb}
\usepackage{mathtools}
\usepackage{enumitem}
\usepackage{multirow,bigdelim}
\usepackage{comment}
\usepackage{framed}
\usepackage{booktabs}
\usepackage{color, colortbl}
\usepackage{tabularx}
\usepackage{soul}
\usepackage{xcolor}
\usepackage{arydshln} 
\usepackage[normalem]{ulem}

\definecolor{purple}{rgb}{0.5,0,1}
\definecolor{dcyan}{rgb}{0.2,0.6,0.5}
\definecolor{light-gray}{gray}{0.95} 
\definecolor{darkgreen}{RGB}{0,140,0}
\definecolor{darkred}{RGB}{200,0,0}
\definecolor{lightgreen}{RGB}{217,231,214}
\definecolor{lightred}{RGB}{255,205,212}
\definecolor{lightyellow}{RGB}{253,242,208}
\definecolor{lightblue}{RGB}{221,231,250}
\definecolor{lightpurple}{RGB}{232,209,255}
\newcommand{\generalname}{Text Modular Networks\xspace}
\newcommand{\generalframework}{TMN framework\xspace}
\newcommand{\generalshortname}{TMNs\xspace}
\newcommand{\name}{\textsc{ModularQA}}

\newcommand{\hotpot}{HotpotQA\xspace}
\newcommand{\drop}{DROP\xspace}
\newcommand{\squad}{SQuAD\xspace}

\newcommand{\roberta}{RoBERTa\xspace}
\newcommand{\bart}{BART\xspace}

\newcommand{\anse}[1]{{\bf \color{teal} #1}}
\newcommand{\quese}[1]{{\color{violet} \textit{#1}}}

\newcommand\T{\rule{0pt}{2.4ex}}

\newcommand{\modqarow}{\rowcolor{violet!15}}
\newcommand{\delimrow}{\rowcolor{gray!10}}

\newcommand{\squadabb}{\ensuremath{\mathcal{S}}}

\newcommand{\mathqabb}{\ensuremath{\mathcal{C}}}

\newcommand{\chain}{\ensuremath{u}}
\newcommand{\context}{\ensuremath{p}}
\newcommand{\ques}{\ensuremath{q}}
\newcommand{\quesset}{\ensuremath{Q}}
\newcommand{\compq}{\ensuremath{qc}}
\newcommand{\ans}{\ensuremath{a}}
\newcommand{\vhintf}{\ensuremath{\Phi}}
\newcommand{\vhint}{\ensuremath{v}}
\newcommand{\submodel}{\ensuremath{t}}

\newcommand{\hint}{\ensuremath{z}}

\newcommand{\decomposer}{next-question generator\xspace}

\newcommand{\colordecomposer}[1]{\colorbox{lightgreen}{#1}}
\newcommand{\colorsubmodel}[1]{\colorbox{lightyellow}{#1}}

\newcommand{\qgen}{sub-task question model\xspace}
\newcommand{\qgens}{sub-task question models\xspace}
\newcommand{\QGen}{Sub-task Question Model\xspace}
\newcommand{\QGens}{Sub-task Question Models\xspace}

\newcommand{\qamodel}{\ensuremath{\mathbb{A}}}
\newcommand{\qgenmodel}{\ensuremath{\mathbb{G}}}
\newcommand{\qdmodel}{\ensuremath{\mathbb{D}}}

\newcommand{\diffc}{\texttt{diff}}
\newcommand{\ifthenc}{\texttt{if\_then}}
\newcommand{\notc}{\texttt{not}}
\newcommand{\textto}{text-to-text\xspace}

\newcommand{\tick}{\textcolor{darkgreen}{\checkmark}}
\newcommand{\notick}{\textcolor{red}{\ensuremath{\times}}}

\title{
\generalname: Learning to Decompose Tasks\\
in the Language of Existing Models \\ 
}
\author{
  Tushar Khot \ \ \ \
    Daniel Khashabi \ \ \ \ 
    Kyle Richardson \\
  \textbf{Peter Clark }\ \ \ 
    \textbf{Ashish Sabharwal}\\
 Allen Institute for AI,
 Seattle, WA, U.S.A.\\
 {\tt \small \{tushark,danielk,kyler,peterc,ashishs\}@allenai.org}\\
}

\date{}

\begin{document}

\maketitle

\begin{abstract}

We propose a general framework called \generalname (\generalshortname) for building interpretable systems that learn to solve complex tasks by decomposing them into simpler ones solvable by existing models.  To ensure solvability of simpler tasks, \generalshortname learn the textual input-output behavior (i.e., \emph{language}) of existing models through their datasets. This differs from prior decomposition-based approaches which, besides being designed specifically for each complex task, produce decompositions independent of existing sub-models. Specifically, we focus on Question Answering (QA) and show how to train a \decomposer to sequentially produce sub-questions targeting appropriate sub-models, without additional human annotation. These sub-questions and answers provide a faithful natural language explanation of the model's reasoning. We use this framework to build \name,\footnote{\label{footnote:code}https://github.com/allenai/modularqa} a system that can answer multi-hop reasoning questions by decomposing them into sub-questions answerable by a neural factoid single-span QA model and a symbolic calculator. Our experiments show that \name\ is more versatile than existing explainable systems for \drop and \hotpot datasets, is more robust than state-of-the-art blackbox (uninterpretable) systems, and generates more understandable and trustworthy explanations compared to prior work.
\end{abstract}

\section{Introduction}
\label{sec:intro}

\begin{figure}
    \centering
        \includegraphics[width=0.92\linewidth,trim=0.2cm 0.5cm 0.2cm 0.5cm]{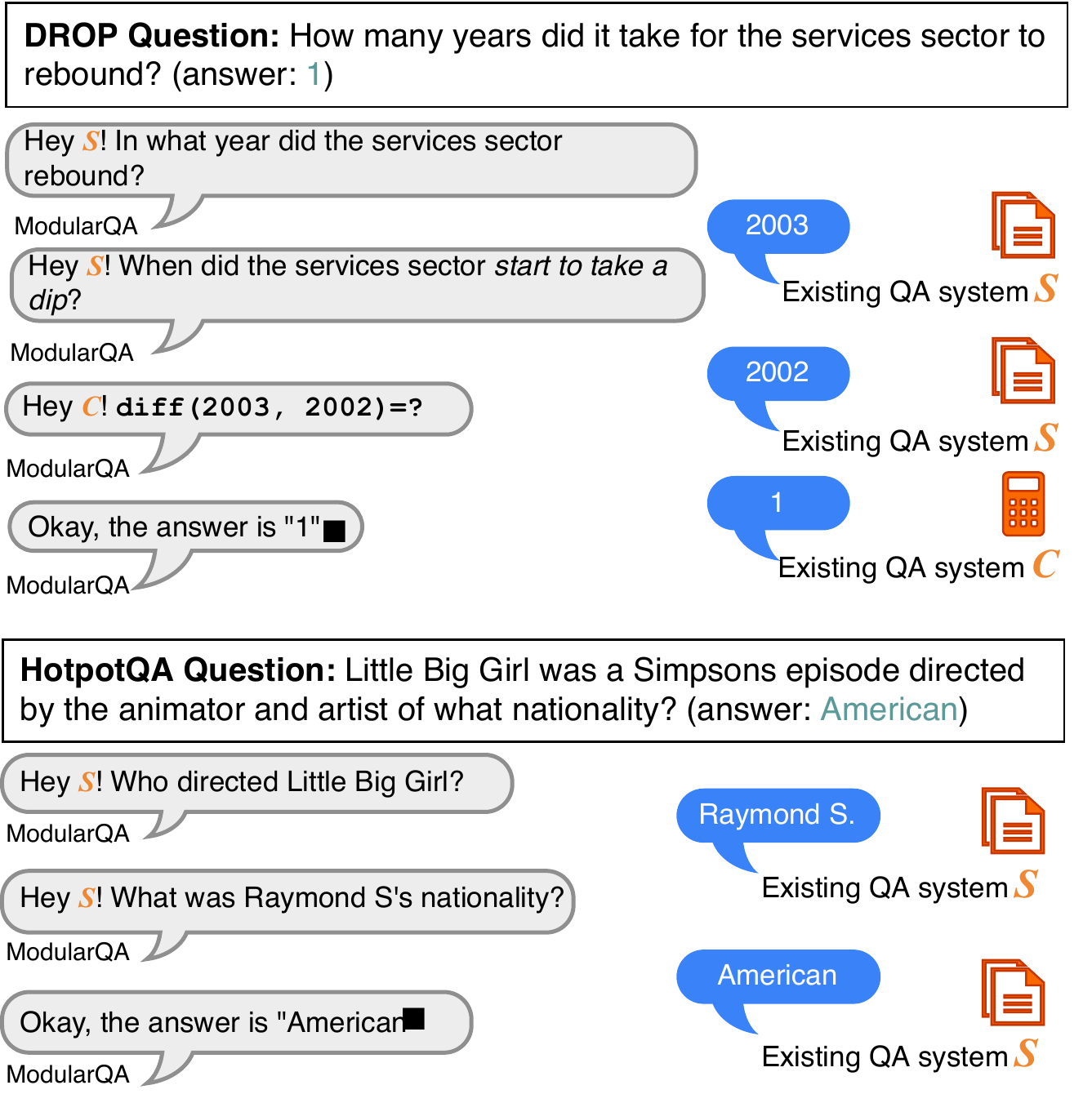}
    \caption{
    \name\ learns to ask sub-questions to existing simple QA models, including a symbolic calculator, to answer a given complex question. Notably, the approach does not rely on annotated decompositions. Despite this, the system learned to add ``start to take a dip'' in the DROP dataset question.}
    \label{fig:intro}
\end{figure}


An intuitive way to solve more complex tasks, such as multi-hop question-answering~\cite{hotpotqa,khashabi2018looking,qasc} and numerical reasoning~\cite{dua2019drop}, would be to decompose them into already solved simpler problems, e.g., single-fact QA~\cite{squad}. Besides allowing reuse of existing simpler models, this approach would yield an \emph{interpretable system} that provides a faithful explanation~\cite{Jacovi2020TowardsFI} of its reasoning as a composition of simpler sub-tasks, as shown in Fig.~\ref{fig:intro}. Motivated by this, we ask the following question:

\emph{Given a set of existing QA models, can one leverage them to  answer complex questions by communicating with these existing models?}

We propose a general framework, \textbf{\generalname (\generalshortname)}, that answers this question by learning to decompose complex questions (of any form) into sub-questions that are answerable by existing QA models---symbolic or neural (henceforth referred to as \colorsubmodel{\emph{sub-models}}).\footnote{\generalshortname, in fact, treat sub-models as blackboxes, and can thus use \emph{any} model or function as a module.} Unlike previous approaches~\cite{talmor2018web,min2019compositional}, the decompositions are not based on splits of the complex questions and aren't built independent of the sub-model. Instead, our framework learns to generate sub-questions in the scope of existing models. For instance, the second sub-question in the DROP dataset example in Fig.~\ref{fig:intro} requires the \emph{introduction of a new phrase}, ``start to take a dip'', which is beyond the scope of standard decomposition approaches. Additionally, the final sub-question targets a \emph{symbolic calculator}, which operates over a different input language.

The core of our \generalframework is a \colordecomposer{\emph{\decomposer}} that sequentially produces the next sub-question to ask as well as an appropriate sub-model for answering it. The resulting sequence of sub-questions and their answers provides a human-interpretable description of the model's \emph{neuro-symbolic} reasoning~\cite{mccarthy1988epistemological,smolensky1988proper},  as illustrated in Fig.~\ref{fig:intro}. Notably, \generalshortname learn to produce these decompositions using only distant supervision, \emph{without} the need for any explicit human annotation.

One of our key insights is that the capabilities of existing sub-models can be captured by training a \textto system to generate the questions in the sub-model's training dataset (e.g., \squad), given appropriate \emph{hints}. In our case, we train a BART model \cite{lewis2019bart} to generate questions given the context, answer, and preferred vocabulary as hints. We then use these \qgens to generate sub-questions (and identify appropriate sub-models) that could lead to the likely intermediate answers extracted for each step of the complex question (``Raymond S.'' and ``American'' in the \hotpot example in Fig.~\ref{fig:intro}). The resulting sub-questions, by virtue of our training, are in the language (i.e., within-scope) of the corresponding sub-models. These sub-question sequences can now be used to \emph{train} the \decomposer to sequentially produce the next sub-question. We use this trained generator, along with existing QA models, to answer complex questions, without the need for any intermediate answers.

We use the \generalframework to develop \textbf{\name}, a modular system that explains its reasoning in natural language, by decomposing complex questions into those answerable by two sub-models: a \emph{neural factoid single-span QA model} and a \emph{symbolic calculator}. \name's implementation\textsuperscript{\ref{footnote:code}} covers multi-hop questions that can be answered using these two sub-models via five classes of reasoning found in existing QA datasets: \emph{composition, conjunction, comparison, difference,} and \emph{complementation}.\footnote{Composition and conjunction questions are also referred to as `bridge' questions. Complementation refers to questions such as `What percentage of X is \emph{not} Y?' \name\ can be easily extended to other reasoning types by defining the corresponding  hints (\S\ref{subsec:extr_hints_modqa}).}

We evaluate \name\ on questions from two datasets, \drop~\cite{dua2019drop} and \hotpot~\cite{hotpotqa}, resulting in the first cross-dataset decomposition-based interpretable QA system.
Despite its interpretability and versatility, \name\ scores only 3.7\% F1 lower than NumNet+V2~\cite{ran2019numnet}, a state-of-the-art \emph{blackbox} model designed for \drop. \name\ even outperforms this blackbox model by 2\% F1 in a limited data setting and demonstrates higher (+7\% F1) robustness~\cite{Gardner2020EvaluatingNM}. \name\ is competitive with and can even outperform task-specific Neural Module Networks~\cite{gupta2020neural,jiang2019self} while producing textual explanations. Further, our human evaluation against a split-point based decomposition model trained on decomposition annotation~\cite{decomprc} for \hotpot finds our explanations to be more trustworthy, understandable, and preferable in 67\%-78\% of the cases.

\paragraph{Contributions.}
(1) \generalname (\generalshortname), a general framework that leverages existing simpler models---neural and symbolic---as blackboxes for answering complex questions. (2) \name,$^{\ref{footnote:code}}$ an interpretable system that learns to automatically decompose multi-hop and discrete reasoning questions. (3) Experiments on \drop and \hotpot demonstrating \name's cross-dataset versatility, robustness, sample efficiency and ability to explain its reasoning in natural language.

\section{Related Work}
\label{sec:related}

Many early QA systems were designed as a combination of distinct modules, often composing outputs of \emph{lower-level} language tasks to solve \emph{higher-level} tasks~\cite{Moldovan2000TheSA,harabagiu2006methods}. However, much of this prior work is limited to pre-determined composition structures~\cite{berant2013semantic, seo2015solving,neelakantan2016learning,roy2018mapping}.

Various \textbf{modular network} architectures have been proposed to exploit compositionality~\cite{rosenbaum2018routing,kirsch2018Modular}.  The closest models to our work are based on \emph{neural module networks} (NMN)~\cite{andreas2016learning} which compose task-specific simple neural modules. We compare against formulations of NMNs for \hotpot~\cite{jiang2019self} and \drop~\cite{gupta2020neural}, both of which target only one dataset and do not reuse existing QA systems. Moreover, they provide attention-based explanations whose interpretability is unclear~\cite{Serrano2019IsAI,Brunner2019OnII,Wiegreffe2019AttentionIN}.

\textbf{Question decomposition} has been pursued before for ComplexWebQuestions~\cite{talmor2018web} and \hotpot. Both approaches~\cite{talmor2018web,decomprc} focus on directly training a model to produce sub-questions using question spans---an approach not suitable for \drop questions (as illustrated in Fig.~\ref{fig:intro}). Our \decomposer overcomes this limitation by generating free-form sub-questions in the language of existing models. \citet{perez2020unsupervised} also use a \textto model to generate sub-questions for \hotpot. However, they generate simpler questions without capturing the requisite reasoning, and hence use them mainly for evidence retrieval. 

BREAK~\cite{break} follows an alternative paradigm of collecting full question decomposition meaning representations (QDMR) annotations. While this can be effective, it relies on costly human annotation that may not generalize to domains with new decomposition operations. Its decompositions are generated in a model-agnostic way and still need QA systems to answer the sub-questions, e.g, high-level QDMR questions such as ``Which is earlier?'' and ``Which is longer?'' would need special systems that can map these to symbolic comparisons. In contrast, \generalshortname\ start with pre-determined models and learn to generate decompositions in their language.

While many \textbf{multi-hop QA} models exist for \hotpot and \drop, these are often equally complex models~\cite{sae,hgn,ran2019numnet} focusing on \emph{just one} of these datasets. Only on \hotpot, where supporting sentences are annotated, can these models also produce post-hoc explanations, but these explanations are often not faithful and shown to be gameable~\cite{Trivedi2020IsMQ}. \generalshortname are able to produce explanations for multiple datasets without needing such annotations, making it more generalizable to future datasets.

\section{\generalname}
\label{sec:TMNs}

\begin{figure}
    \centering
    \includegraphics[scale=0.55,trim=0cm 0.0cm 0cm 0.0cm]{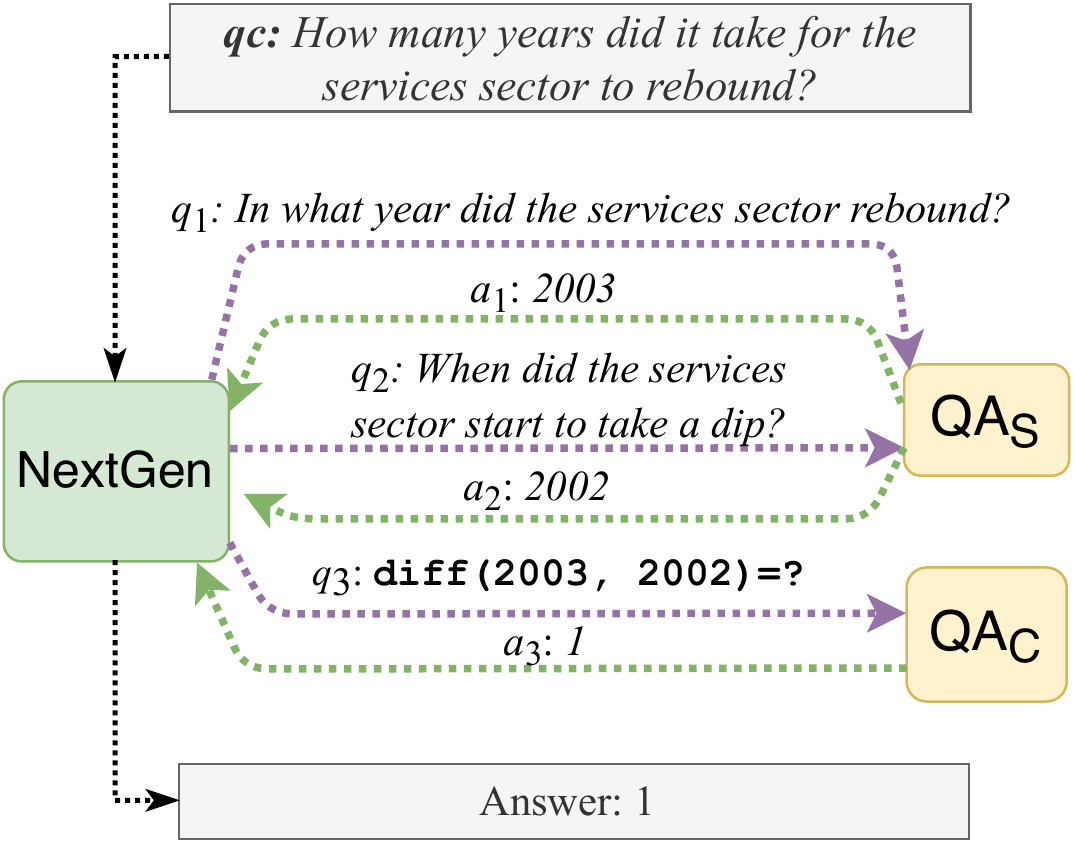}
    \caption{A sample inference on a \drop question using \generalshortname with the text-to-text interactions between the \colorbox{lightgreen}{\decomposer \qdmodel} and \colorbox{lightyellow}{existing QA models \qamodel}.}
    \label{fig:toy_inf}
\end{figure}

\generalshortname are a family of architectures consisting of \emph{modules} that communicate through \emph{language} learned from these modules, to accomplish a certain goal (e.g., answering a question). Figure~\ref{fig:toy_inf} illustrates this general idea in the context of answering a \drop question. The core of our system is a \colorbox{lightgreen}{\emph{\decomposer} \qdmodel}, a component in charge of generating and distributing sub-tasks among \colorbox{lightyellow}{sub-models $\qamodel_{s}$}. 
The system alternates between using \qdmodel\ to produce the next question (\emph{NextGen}) and using the corresponding sub-model to answer this question. Formally, solving a complex question $\compq$ is an alternating process between the following two steps: \\
\noindent
{\it Generate the next question $\ques_{i}$ for submodel $\submodel_i$:} \\ 
\hspace*{3mm} $\langle \submodel_i, \ques_{i} \rangle = \qdmodel(\compq, \ques_1, \ans_1, \hdots, \ques_{i-1}, \ans_{i-1})$ \\ 
{\it Find answer $\ans_i$ by posing $\ques_{i}$ to submodel $\submodel_i$:} \\
\hspace*{6mm} $\ans_i \hspace*{4mm} = \qamodel_{\submodel_i}(\ques_{i}, p) $\\
\noindent where $\ques_i$ is the $i^{th}$ generated sub-question and $\ans_i$ is the answer produced by a sub-model $\submodel_i$ based on a given context paragraph $p$. 
This simple iterative process ends when $\ques_{i+1}$ equals a special end-of-sequence symbol (denoted throughout as \texttt{[EOQ]}) with the final output answer $\ans_{i}$. 

\paragraph{Building a Text Modular Network.}
The key challenge in building a \generalname\ is developing the \decomposer model. Training this model requires a next-question prediction dataset where each example is a step in the iterative progression of sub-question generation. For example, the second step in Fig.~\ref{fig:toy_inf} is:

\newcolumntype{L}{>{\centering\arraybackslash}m{6cm}}

\noindent \begin{tabular}{ccl}
     \parbox[t]{0mm}{\multirow{5}{*}{\rotatebox[origin=c]{90}{\textbf{In}}}} & \rdelim\}{5}{0mm}[] & 
     \parbox[t]{0mm}{\multirow{5}{*}{\parbox{6.7cm}{ \textit{$\compq$: How many years did it take for the services sector to rebound? \\
 $\ques_1$: In what year did the services sector rebound? \\
 $\ans_1$: 2003  }}}}\\ 
     & & \\
     & & \\
     & & \\
     & & \\
     \parbox[t]{0mm}{\multirow{2}{*}{\rotatebox[origin=c]{90}{\textbf{Out}}}}  & \rdelim\{{2}{0mm}[] &
     \parbox[t]{0mm}{\multirow{2}{*}{\parbox{6.7cm}{ \textit{
     $\langle \submodel_2, \ques_{2} \rangle$=\textit{
 $\langle$\squad, ``When did the services sector start to take a dip?''$\rangle$
     }  }}}}  \\
     & & \\ 
 \end{tabular}

While it may be possible to collect task-specific datasets or design a task-specific \decomposer~\cite{decomprc,talmor2018web}, our goal is to build a framework that can be easily extended to new complex QA tasks  reusing existing QA sub-models. To achieve this, we 
present a {\it general} framework to {\it generate the next-question training dataset} by: (1) Modeling \emph{the language} of sub-models; (2) Building decompositions in the language of these sub-models using minimal distant supervision \emph{hints}.

\begin{figure*}[ht]
    \centering
    \includegraphics[scale=0.85,trim=0.5cm 0.5cm 0.5cm 0.5cm]{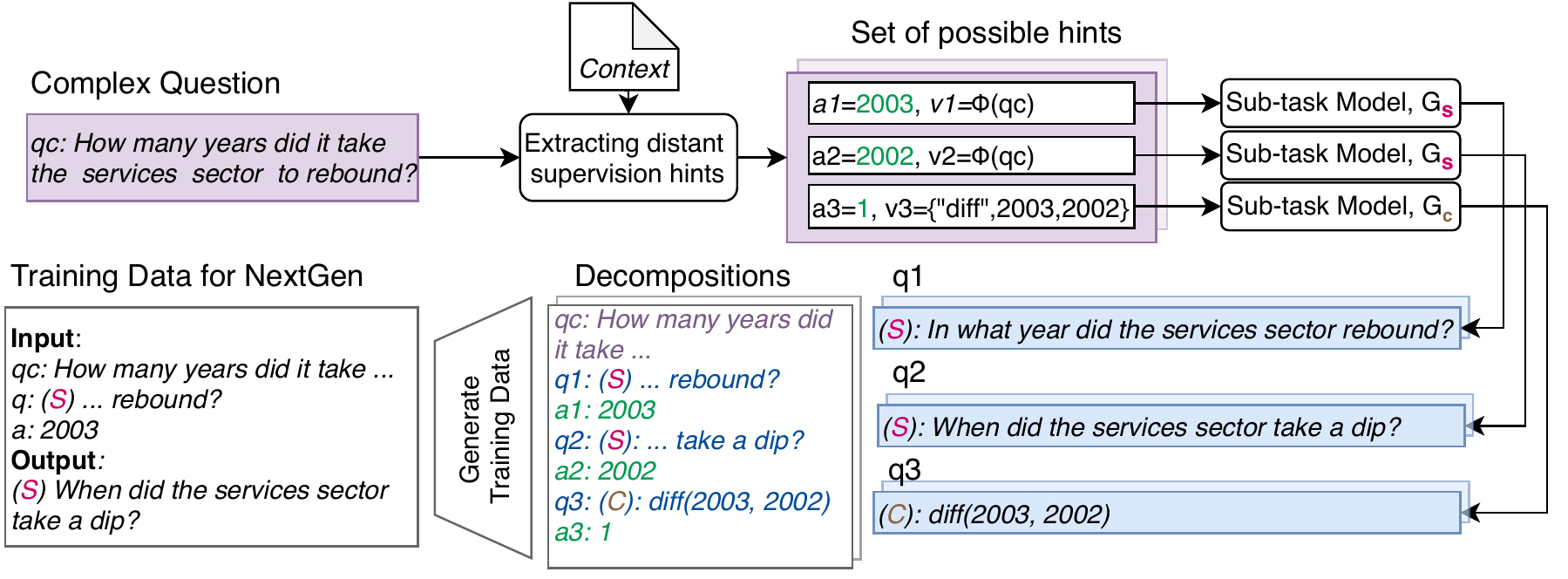}
    \caption{The overall flow of building the training data for the \decomposer, given a complex question. }
    \label{fig:data_flow}
\end{figure*}


\subsection{Modeling QA Sub-Models}
\label{subsec:lang_mod}
To ensure the sub-questions are answerable by existing sub-models, we train a \textto \emph{\qgen} on the original sub-task to generate a plausible $\ques_i$ conditioned on hints, e.g., a \bart model trained on \squad to generate a question given the answer. We can view this utility as characterizing the question language of the sub-model. For example, such a model trained on the \squad dataset would produce factoid questions---the space of questions answerable by a model trained on this dataset.

While an unconditional text generation model can also capture the space of questions, it can generate a large number of possibly valid questions, making it hard to effectively train or use such a model. Instead, we scope the problem down to conditional text generation of questions given hints $\hint$.   Specifically, we use the context \context, answer \ans\ and question vocabulary \vhint\ as input conditions to train a question generator model $\qgenmodel: \hint \rightarrow \ques  $ where $\hint=\langle \context, \ans, \vhint\rangle$.
 Such a generator, $\qgenmodel_\squadabb$, produces the first two sub-questions in the example in Fig.~\ref{fig:decomp}, when using \ans=2003 (or 2002, resp.) and \vhint=$\vhintf(\compq)$=\{``service'', ``sector'', ``year'', ``rebound''\} as hints.

\subsection{Training Decompositions via Distant Supervision}
\label{subsec:extr_hints}

To generate training decompositions for a complex question using a \qgen, we  extract distant-supervision hints $\hint$ corresponding to each reasoning step. This is akin to the distant supervision approaches used to extract logical forms in semantic parsing~\cite{liang2013learning,berant2013semantic} and the intermediate entities in a reasoning chain~\cite{gupta2020neural,jiang2019self}.

\begin{figure}[ht]
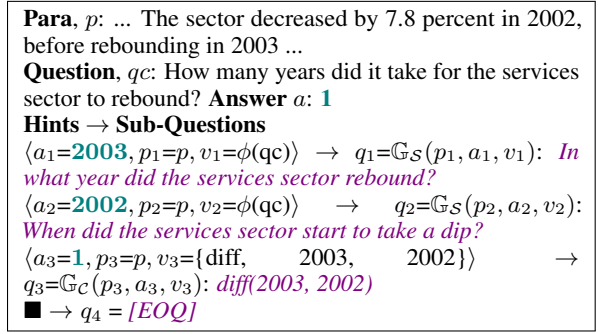

    \centering
    \fbox{
    \parbox{0.95\linewidth}{
        \small
        \textbf{Para}, \context: ... The sector decreased by 7.8 percent in 2002, before rebounding in 2003 ...\\
        \textbf{Question}, \compq: How many years did it take for the services sector to rebound? 
        \textbf{Answer} \ans: \anse{1}\\
        \textbf{Hints $\rightarrow$ Sub-Questions}\\
        $\langle \ans_1$=$\anse{2003}, \context_1$=$\context, \vhint_1$=$\phi$(qc)$ \rangle \rightarrow
        \ques_1$=$\qgenmodel_{\squadabb}(\context_1, \ans_1, \vhint_1)$:  \quese{In what year did the services sector rebound?} \\
        $\langle \ans_2$=$\anse{2002}, \context_2$=$\context, \vhint_2$=$\phi$(qc)$  \rangle \rightarrow
        \ques_2$=$\qgenmodel_{\squadabb}(\context_2, \ans_2, \vhint_2)$: \quese{When did the services sector start to take a dip?} \\
        $\langle \ans_3$=$\anse{1}, \context_3$=$\context, \vhint_3$=\{diff, 2003, 2002\}$ \rangle \rightarrow
        \ques_3$=$\qgenmodel_{\mathqabb}(\context_3, \ans_3, \vhint_3)$: \quese{diff(2003, 2002)} \\
        $\blacksquare \rightarrow \ques_4$ = \quese{[EOQ]}
    }
    }
    \caption{An example decomposition generated for a \drop\ example using hints and sub-question generators \qgenmodel. $\phi$(qc) indicates words in the input question qc.
    }
    \label{fig:decomp}
    \end{figure}

In our running \drop example, under the definition of $\hint = \langle \context, \ans, \vhint \rangle$, we would need to provide the context, answer, and question vocabulary for each reasoning step. We can derive intermediate answers by finding the two numbers whose difference is the final answer (see Fig.~\ref{fig:decomp}). We can use words from the input question as vocabulary hints.\footnote{As mentioned before, these are soft hints and the model can be trained to handle noise in these hints.} 

As shown in Fig.~\ref{fig:decomp}, we generate the training sub-questions in the language of appropriate systems for each step $i$ using the question generation model $\qgenmodel_{\submodel_i}$: $\ques_i = \qgenmodel_{\submodel_i}(\hint_i)$ where $\hint_i=\langle \context_i, \ans_i, \vhint_i \rangle$ and the model $\submodel_i$ is determined by the answer type (or can be a hint too). 

Note that our framework does not depend on the specific choice of \hint. Our key idea is to train the \qgen conditioned on the same \hint\ that we can provide for the complex task. The hint \hint\ could be very general (just the context) or very specific (exact vocabulary of the question), trading off the ease of extracting hints with the quality of the generated decomposition. Similarly, these hints don't have to be 100\% accurate as they are only used to build the training data and play no role during inference.

Finally, we convert the decompositions into training data for the \decomposer. For each question $q_i$ generated using the \qgen $\qgenmodel_{\submodel_i}$, we create the training example:

\parbox[c]{0.95\linewidth}{
 \textit{
 Input: $\compq, \ques_1, \ans_1, \ldots, \ques_{i-1}, \ans_{i-1}$\\
 Output: $\langle \submodel_i, \ques_i \rangle$
 }
}

\paragraph{Training Data Generation Summary.}
Fig.~\ref{fig:data_flow} illustrates the complete process for generating the training data for the \decomposer. For each complex question, we extract a set of possible hints for each potential reasoning chain (e.g., all number pairs that lead to the final answer). For each step, we use the corresponding \qgens\ to generate potential sub-questions that lead to the expected answer. Finally we use these generated sub-question decompositions as the training data for the \decomposer model.

\section{\name\ System}
\label{sec:system}

We next describe a specific instantiation of the Text Modular Network: \name\ -- a new QA system that works across \hotpot and \drop. To handle these datasets, we first introduce the two QA sub-models(\S\ref{subsec:lang_mod_impl}), the \qgens for these models(\S\ref{subsec:qgen}), our approach to build training data (\S\ref{subsec:extr_hints_modqa}), and the inference procedure used for question-answering(\S\ref{subsec:inference}).

\subsection{QA Sub-Models, $\qamodel$}
\label{subsec:lang_mod_impl}
We use two QA models with broad coverage on the two datasets:\\
%
    \textbf{\squad model, $\qamodel_\squadabb$}, A \roberta-Large model trained on the entire \squad 2.0 dataset including the no-answer questions; and\\
    \textbf{Math calculator model, $\qamodel_\mathqabb$,} a symbolic Python program that can perform key operations needed for \drop and \hotpot, namely:\\
    \indent \diffc(X, Y, Z) that computes the difference between X and Y in unit Z (days/months/years); \\
    \indent \notc(X) that computes the complement \% of X, i.e., 100 - X;\\
    \indent \ifthenc(X <op> Y, Z, W) that returns Z if X <op> Y is true, otherwise returns W.

\subsection{\QGens, $\qgenmodel$}
\label{subsec:qgen}
We define two \qgens corresponding to each of our QA sub-models.

\paragraph{\squad \QGen, $\qgenmodel_\squadabb$.} 
We train a \bart-Large model on the answerable subset of \squad 2.0 to build our \qgen for \squad. We use the gold paragraph and answer from the dataset as the input context and answer. For the estimated question vocabulary, we select essential words\footnote{
$\vhintf(q)$ = Non-stopword tokens with pos tags $\in$ \{NOUN, VERB, NUM, PROPN, ADJ, RB\}
} from the gold questions (referred as the function $\vhintf$) with additional irrelevant words sampled from other questions.\footnote{More details in Appendix~\ref{app:hyper}}

To train the \textto $\mathrm{\bart}_\squadabb$ model, we use a simple concatenation of the passage, vocabulary, and answer (with markers such as ``H:'' and ``A:'' to indicate each field) as the input sequence and the question as the output sequence. While a constrained-decoding approach~\cite{hokamp2017lexically,hu2019improved} could be used here to further promote the use of the vocabulary hints, this simple approach was effective and more generally applicable to other hints in our use-case. 

Once this model is trained, we use it with nucleus sampling~\cite{nucleus} to generate $k$ sub-questions, $\quesset$, and filter out those that lead an incorrect or no answer using $\qamodel_\squadabb$:

$\qgenmodel_\squadabb(\context, \ans, \vhint)  =  \{\ques \in \quesset\mid\textrm{overlaps}(\qamodel_\squadabb(\context, \ques), \ans)\}$

\paragraph{Math \QGen, $\qgenmodel_\mathqabb$.}
Given the symbolic nature of this solver, rather than training a neural generator, we simply generate all possible numeric questions given the context. Similar to $\qgenmodel_\squadabb$, we first generate potential questions $Q$ and then filter down to those that lead to the expected answer using $\qamodel_\mathqabb$:

$\qgenmodel_\mathqabb(\context, \ans, \vhint)  = \{\ques \in Q \mid \qamodel_\mathqabb(\context, \ques) = \ans\}
$


\begin{table*}[t]
    \centering
    \small
    \resizebox{\textwidth}{!}{
    \begin{tabular}{p{1.2\textwidth}}
        \multicolumn{1}{c}{\large \drop}\\
        \midrule
        {\normalsize \underline{Example 1}}: \textbf{How many days passed between the Sendling Christmas Day Massacre and the Battle of Aidenbach? }\\
          >>   Q: \quese{When was the Battle of Aidenbach?} A: 8 January 1706 Q: \quese{When was the Sendling Christmas Massacre?} A: 25 December 1705 Q: \quese{\diffc(8 January 1706, 25 December 1705, days)} A: \anse{14}\\
        {\normalsize \underline{Example 2}}: \textbf{ Which ancestral group is smaller: Irish or Italian? } \\
         >>  Q: \quese{How many of the group were Irish?} A: 12.2 Q: \quese{How many Italian were there in the group?} A: 6.1 Q: \quese{\ifthenc(12.2 < 6.1, Irish, Italian)} A: \anse{Italian} \\
         {\normalsize \underline{Example 3}}: \textbf{How many percent of the national population does not live in Bangkok? } \\
          >>  Q: \quese{What percent of the national population lives in Bangkok?} A: 12.6 Q: \quese{\notc(12.6)} A: \anse{87.4} \\
         \\
        \multicolumn{1}{c}{\large \hotpot}\\
        \midrule
        {\normalsize \underline{Example 4}}: \textbf{12 Years a Slave starred what British actor born 10 July 1977) }\\
         >> Q: \quese{Who stars in 12 Years a Slave?} A: Chiwetel Ejiofor Q:  \quese{Who is the British actor born 10 July 1977?} A: \anse{Chiwetel Umeadi Ejiofor}   \\
        {\normalsize \underline{Example 5}}: \textbf{How many children's books has the writer of the sitcom Maid Marian and her Merry Men written ?}   \\
          >> Q:  \quese{What writer was on Maid Marian and her Merry Men?} A: Tony Robinson Q:  \quese{How many children's books has Tony Robinson written?} A: \anse{sixteen} \\
         {\normalsize \underline{Example 6}}: \textbf{Did Holland's Magazine and Moondance both begin in 1996?  }\\
          >>  Q: \quese{When did Holland's Magazine begin?} A: \anse{1876} Q: \quese{When did Moondance begin?} A: \anse{1996} Q: \quese{\ifthenc(1876$\neq$1996, no, yes)} A: \anse{no} \\
          \bottomrule
    \end{tabular}
    }
    \caption{
    Sample Reasoning Explanations generated by \name. Note that the system learns to generate such explanations without relying on manually designed rules such as ``smaller'' $\Rightarrow x < y$.
    }
    \label{tab:sample_decomp}
\end{table*}


\subsection{Generating Training Decompositions}
\label{subsec:extr_hints_modqa}

We broadly identify five classes of questions in \hotpot and \drop dataset that can be answered using our two models.\footnote{Other questions require a QA model that can return multiple answers or a Boolean QA model, as discussed in \S\ref{sec:discussion}.} These question classes, how they are identified and how we extract hints for each question type is described next. Note that similar rules for extracting distant supervision hints have been used by prior work for \drop~\cite{gupta2020neural} and \hotpot~\cite{jiang2019self} too.

1.~\textbf{Difference} (\textit{How many days before X did Y happen?}): We identify these questions based on the presence of term indicating a measurement :``how many'' and terms indicating difference such as ``shorter'', ``more'', ``days between'', etc. Also we check for two dates or numbers in the context such that their difference (in all units) can lead to the final answer. If these conditions are satisfied, for every pair $n_1$, $n_2$ where the difference (in units $u$) can lead to the final answer, we generate the hints:\\
  \indent $\context_1 = \context; \ans_1=n_1; \vhint_1=\vhintf(\compq)$ \\
  \indent $\context_2 = \context; \ans_2=n_2;  \vhint_2=\vhintf(\compq)$\\
  \indent $\context_3 = \varepsilon; \ans_3=\ans; \vhint_3=[\diffc, n_1, n_2, u]$\\
where $\varepsilon$ refers to the empty string.

2.~\textbf{Comparison} (\textit{Which event happened before: X or Y?}): We identify the two entities $e_1$ and $e_2$ in such questions and find dates/numbers that are mentioned in documents. For every $n_1$, $n_2$ number/date mentioned close to $e_1$ and $e_2$ respectively, we create the hints:\\
  $\context_1 = p; \ans_1=n_1; \vhint_1=\vhintf(\compq) \setminus e_2$ \\
  $\context_2 = p; \ans_2=n_2;  \vhint_2=\vhintf(\compq) \setminus e_1$ \\
  $\context_3 = \varepsilon; \ans_3=\ans; \vhint_3=[\ifthenc, n_1, n_2, e_1, e_2]$\\
The final set of hints are for use by the calculator generator to create the questions: $\ifthenc(n_1 > n_2, e_1, e_2)$ and $\ifthenc(n_1 < n_2, e_1, e_2)$.

3.~\textbf{Complementation} (\textit{What percent is not X?}): We identify these questions mainly based on the presence of ``.* not .*'' in the question and a number $n_1$ such that the $\ans=100-n_1$. The hints are:\\
  \indent $\context_1 = \context; \ans_1=n_1; \vhint_1=\vhintf(\compq)$ \\
  \indent $ \context_2 = \varepsilon; \ans_2=\ans; \vhint_2=[\notc, n_1]$

4.~\textbf{Composition} (\textit{Where was 44th President born?}): 
For such questions(only present in \hotpot), we need to first find an intermediate entity $e_1$ that would be the answer to a sub-question in \compq\ (e.g. Who is the 44th President?). This intermediate entity is used by the second sub-question to get the final answer. Given the two gold paras $d_1$ and $d_2$, where $d_2$ contains the answer, we use the mention of $d_2$'s title in $d_1$ as the intermediate entity.\footnote{If not found, we ignore such questions.} While we could use the entire complex question vocabulary to create hints, we can reduce some noise by removing terms that appear exclusively in the other document. So the final hints are:\\
  \indent $\context_1  = d_1; \ans_1=e_1; \hint_1=\zeta(\compq, d_1, d_2)$ \\
  \indent $\context_2 = d_2; \ans_2=\ans; \hint_2=\zeta(\compq, d_2, d_1) + e_1$\\
where $\zeta(\ques, d_1, d_2)$ indicates the terms in $\vhintf(\ques)$ that appear in $d_2$ but not in $d_1$.\footnote{We use the same for comparison questions in \hotpot.}

5.~\textbf{Conjunction} (\textit{Who acted as X and directed Y?}): These class of questions do not have any intermediate entity but have two sub-questions with the same answer e.g. ``Who is a politician and an actor?''. If the answer appears in both supporting paragraphs, we assume that it is a conjunction question. The hints for such questions are:\\
  \indent $\context_1 = d_1; \ans_1=\ans; \hint_1=\zeta(\compq, d_1, d_2)$ \\
  \indent $\context_2 = d_2; \ans_2=\ans; \hint_2=\zeta(\compq, d_2, d_1)$

While decomposition datasets such as BREAK could be used to obtain more direct supervision for these hints, we focus here on the broader feasibility of distant supervision. We observe that our current approach generates hints for 89\% of the questions and can find decompositions that lead to the gold answer for 50\% of them. So while the hints cannot be used directly to produce decompositions, the \decomposer is able to generalize from these examples to generate decompositions for all questions with 81\% of them leading to the gold answer. App.~\ref{app:intera} provides more details and example of hints for each question class.

As described earlier, given these input hints and our \qgens, we can generate the sub-question for each step and the appropriate sub-model (based on the model that produced this question).
 We use nucleus sampling to sample 5 questions for each reasoning step. To improve the training data quality, we also filter out potentially noisy decompositions.\footnote{if an intermediate answer is unused or vocabulary of question chain is too different from the input question. See Appendix~\ref{app:decomp} for more details.} 
 We train a \bart-Large model, our \decomposer, on this training data to produce the next question given the complex question and previous question-answer pairs.

\subsection{Inference}
\label{subsec:inference}
We use best-first search \cite{dijkstra1959note} to find the best decomposition chain and use the answer produced at the end of the chain as our predicted answer. We sample $n_0$ sub-questions from the \decomposer using nucleus sampling. Each question is then answered by the appropriate QA sub-model (defined by the prefix in the question). This partial chain is again passed to the \decomposer to generate the next $n_1$ sub-questions, and so on.\footnote{To enable early exploration, we use exponential decay on the number of generated questions: $n_i = 10/2^i$.} A chain is considered complete when the \decomposer outputs the end-of-chain marker \texttt{[EOQ]}. 

We define a scoring function that scores each partial chain $\chain$ based on the new words introduced in the sub-questions compared to the input question.\footnote{$\theta(\chain)$ = \#new words/\#words in input question} For a complete chain, we additionally add the score from a \roberta model trained on randomly sampled chains (chains that lead to the correct answer are labeled as positive). Concretely, we use the negative class score from this classifier, $\delta(\chain)$, to compute the final chain score as $\theta(\chain) + \lambda \delta(\chain)$, i.e., lower is better.\footnote{For more details, refer to App.~\ref{app:inference}.}

\section{Experiments}

To evaluate our modular approach, we use two datasets, \drop and \hotpot, that contain questions answerable using a \squad model and a math calculator. We identify 14.4K training questions in DROP that are within the scope of our system,\footnote{See App.~\ref{app:intera} for how this subset is automatically identified.} which forms 18.7\% of the dataset.\footnote{Previous modular systems~\cite{gupta2020neural} have targeted even smaller subsets to develop modular approaches.} We similarly select 2973 Dev questions (from 9536), and split them into 601 Dev and 2371 Test questions. 

We evaluate our system on the entire \hotpot dataset. Since the test set is blind, we split the Dev set (7405 qns.) into 1481 Dev and 5924 Test questions. For training, we only use 17\% of the training dataset containing 15661 questions categorized as ``hard'' by \hotpot authors.\footnote{Increasing the training set didn't affect performance.}

\begin{table*}[ht]
    \centering
    \small
    \setlength{\tabcolsep}{12pt}
    \begin{tabular}{lcccccccc}
       & \multicolumn{4}{c}{\drop F1} & \ \ \ & \multicolumn{3}{c}{\hotpot F1} \\ 
        \cmidrule(lr){2-5} \cmidrule(lr){7-9}
        & All & Diff & Comp & Cmpl & & All & Br & Comp \\
        \toprule
     \delimrow \multicolumn{9}{c}{Interpretable Cross-Dataset Models (\S\ref{subsec:interpretable-versatile})}\\
    \modqarow \name &  87.9 & 85.2 & 81.0 & 96.6 & & 61.8 & 64.9 & 49.2 \\
    \addlinespace[2pt]
     WordOverlap  & 80.5 & 82.5 & 58.3 & 95.8 & & 57.5 & 61.7 & 40.5 \\
     Greedy & 60.2 & 52.2 & 52.9 & 76.3 & & 42.4 & 44.8 & 33.0 \\
     \delimrow \multicolumn{9}{c}{Limited Versatility (\S\ref{subsec:not-versatile})}\\
    NMN-D$\dagger$\; &  \phantom{0}79.1* & -- & -- & -- & & -- & -- & --  \\
    SNMN$\dagger$  & -- & -- & -- & -- & & 63.1 & 63.7 & 60.1 \\ 
    DecompRC$\dagger$  & -- & -- & -- & -- & & 70.3 & 72.1 & 63.4 \\ 
    \delimrow     \multicolumn{9}{c}{Limited Interpretability (\S\ref{subsec:not-interpretable})}\\
        NumNet+V2 &   91.6 & 86.5 & 94.5 & 95.5 & & -- & -- & -- \\
        Quark & -- &  -- & -- & -- & & 75.5 & 78.1 & 64.9 \\
        \bottomrule
    \end{tabular}
    \caption{F1 scores on the \drop and \hotpot questions and the individual classes: Difference(Diff), Comparison(Comp), Complementation(Cmpl) and Bridge(Br). TOP: Comparison to variations of \name\ that work across datasets. MIDDLE: Comparison to \emph{targeted} interpretable systems. BOTTOM: Comparison to \emph{targeted blackbox} systems. \name\ is competitive with previous approaches on \drop and mainly lags behind systems on \hotpot that are able to exploit artifacts.
    }
    \label{tab:results}
\end{table*}


\subsection{Explanation and Interpretability}

A key aspect distinguishing \name\ is that it can explain its reasoning in a human-interpretable fashion, in the form of simpler sub-questions it creates via decomposition. Table~\ref{tab:sample_decomp} illustrates six sample reasoning explanations; the question context and sub-models are omitted for brevity. We see that \name\ is able to take oddly phrased questions to create clean sub-questions (example 4), handle yes/no questions (example 6), recognize the unit of comparison (example 1), and map the phrase ``smaller" to the appropriate direction of comparison without any manual rules (example 2).

Analyzing such explanations for 40 Dev questions (20 from each dataset), we found that among the 28 questions \name\ answered correctly, it produced a valid reasoning chain in as many as 93\% of the cases, attesting to its strong ability to provide understandable explanations.

To further assess the human readability of \name's explanations, we compared them with those produced by DecompRC~\cite{decomprc}, the only decomposition-based system for the considered datasets. We identified 155 questions that are within the scope of \name\ and for which both systems produce a decomposition.\footnote{DecompRC failed to produce chains on 6x more questions than our system. See App.~\ref{app:human_eval} for details on how these questions were selected and how they were normalized.} We then asked crowdworkers on Amazon Mechanical Turk to annotate them along three dimensions: (1) given the two explanations, which system's answer do they \textbf{trust} more; (2) which system's explanation do they \textbf{understand} better; and (3) which system's explanation do they generally \textbf{prefer}.

\begin{table}[ht]
    \centering
    \small
    \setlength{\tabcolsep}{4pt}
    \begin{tabular}{lccc}
         & Trust & Understand & Prefer \\
         \midrule
         DecompRC & \phantom{0}50 (33\%) & \phantom{0}34 (22\%) & \phantom{0}49 (32\%) \\
         \modqarow \name & 105 (67\%) & 121 (78\%)& 106 (68\%)\\
    \end{tabular}
    \caption{Human evaluation of the explanation quality. Across all dimensions, crowdsource workers preferred the explanations of \name\ over DecompRC.}
    \label{tab:human_eval}
\end{table}


Table ~\ref{tab:human_eval} summarizes the aggregate statistic of the majority labels, with 5 annotations per question. Crowdworkers understood \name's natural language explanations better in 78\% of the cases, trusted more that it pointed to the correct answer, and generally preferred its explanations.

\subsection{Interpretable Cross-Dataset Models}
\label{subsec:interpretable-versatile}

With \name\ being the first interpretable model for \drop and \hotpot, there were no comparable existing cross-dataset systems. We instead consider two baselines obtained by modifying \name: (1) only the word-overlap based scoring function $\theta(\chain)$ for chains (no \roberta classifier); and (2) greedy inference, i.e., use the most likely question at each step (no search).

As shown in Table~\ref{tab:results} (top rows), \name\ outperforms the purely word-overlap based approach by 7pts F1 on \drop and 4pts on \hotpot. A simple coverage-based decomposition is thus not as effective, although \hotpot suffers less because of decompositions being explicit in it.\footnote{Recall that our word-overlap based score penalizes missed question words and words introduced during decomposition.} Performance drops much more heavily (18pts on \drop and 19pts on \hotpot) when we do not employ search at all. This is primarily because the optimal sub-question can often be unanswerable by the intended sub-model while an alternate decomposition may lead to the right answer.

\subsection{Comparison to Dataset-Specific Models}
\label{subsec:not-versatile}

To assess the price \name\ pays for being versatile, we compare it to three interpretable systems that target a particular dataset. Two are Neural Module Networks, with modules designed specifically for a subset of \drop (referred to as NMN-D)~\cite{gupta2020neural} and for \hotpot (referred to as SNMN)~\cite{jiang2019self}. The third is DecompRC, whose split-based decomposition, human annotations, as well as answer composition algorithm was specifically designed for \hotpot. 

As seen in Table~\ref{tab:results} (middle rows), \name\ actually substantially outperforms the \drop model NMN-D while being able to produce textual explanations (rather than attention visualization).\footnote{Since NMN-D focuses on a different subset, we report its score on the shared subset, on which \name\ achieves an F1 score of 92.5 (not shown in the table)} On the \hotpot dataset, \name\ is comparable to S-NMN but underperforms compared to DecompRC. Note that DecompRC can choose to answer some questions using single-hop reasoning and potentially exploit many artifacts in this dataset~\cite{min2019compositional,Trivedi2020IsMQ}.

\subsection{Comparison to Black-Box Models}
\label{subsec:not-interpretable}

To assess the price \name\ pays for being interpretable, we compare it to two state-of-the-art black-box systems that not only lack interpretability but are also targeted towards specific datasets: NumNet+V2~\cite{ran2019numnet} for \drop and Quark~\cite{groeneveld2020simple} for \hotpot. Since we use the \squad QA system in our model, we first fine-tune the LM in both of these systems on the \squad dataset, and then train them on the same datasets as \name.

As seen in Table~\ref{tab:results} (bottom rows), we are competitive with the state-of-the-art model on \drop but underperform compared to the Quark system. Note that Quark relies on supporting fact annotation and trains a single end-to-end QA model, thereby being more likely to exploit dataset artifacts.

Upon analyzing \name's errors (defined as questions with F1 score under 0.5) on \hotpot, we found 65\% of the errors arise from intermediate questions having multiple or yes/no answers. These are not handled by modules in our current implementation, suggesting a path for improvement.

We also analyzed the errors on the DROP dev set and identified question decomposition (53.3\%) and QA models (33.33\%) as the main sources of error.\footnote{Remaining errors are due to dataset and scope issues.} Within question decomposition, the key cause of error is higher \roberta score for an incorrect decomposition (50\% of errors). Both the \squad and Math QA models were responsible for errors, with the latter erring only due to out-of-scope formats (e.g., date ranges 1693-99). Appendix~\ref{app:drop_err} provides more details.

\subsection{Additional Benefits of \generalshortname}
The last set of experiments support two distinct benefits (besides interpretability) of our approach even against state-of-the-art black-box models.

\paragraph{Higher Robustness.}
We evaluate on the \drop \emph{contrast set}~\cite{Gardner2020EvaluatingNM}, a suite of test-only examples created for assessing robustness via minimally perturbed examples. On the 239 (out of 947) questions that are within our scope using the same logic as before, we find that \name\ outperforms NumNet+V2 by 7\%-10\%:
\begin{center}
 \small
    \begin{tabular}[t]{l|cc}
    Contrast Test & EM & F1\\
    \modqarow \name &  \textbf{55.7} &\textbf{63.3} \\
    NumNet+V2    &  45.2 & 56.2
    \end{tabular}
\end{center}

\paragraph{Learning with Less Data.}
We next evaluate the sample efficiency of \name\ by considering training sets of 3 different sizes: 100\%, 60\%, and 20\% (14448, 8782, and 2596 questions, resp.) of the training questions selected for \drop.\footnote{For simplicity, we train 
\name\ on the DROP questions only here. To obtain sufficient examples, we increase the number of questions sampled for each decomposition step. See App.~\ref{app:less_data} for more details.} As shown below, the gap (in F1 score) between \name\ and the state-of-the-art model steadily shrinks, and \name\ even outperforms it when both are trained on 20\% of the data.

\begin{center}
 \small
    \begin{tabular}[t]{l|ccc}
    Portion of Train set & 100\% & 60\% & 20\%\\
    \modqarow \name & 87.8 &  \textbf{89.3} & \textbf{87.0}  \\
    NumNet+V2    & \textbf{91.6} & 88.3 & 85.4 
    \end{tabular}
\end{center}


\section{Conclusion \& Future Work}
\label{sec:discussion}
We introduced \emph{\generalname}, which provide a general-purpose framework that casts complex tasks as \emph{textual} interaction between existing, simpler QA modules. Based on this conceptual framework, we built \name, an instantiation of \generalshortname that can perform multi-hop and discrete numeric reasoning. Empirically, \name\ is on-par with other modular approaches (which are dataset-specific) and outperforms a state-of-the-art model in a limited data setting and on expert-generated perturbations. Importantly, \name\ provides easy-to-interpret explanations of its reasoning. It is the first system that decomposes \drop questions into textual sub-questions and can be applied to both \drop and \hotpot.

Extending this model to more question classes such as counting (``How many touchdowns were scored by X?'') and Boolean conjunction (``Are both X and Y musicians?'') are interesting avenues for future work. To handle the former class, the first challenge is building models that can return a list of answers---a relatively unexplored task until recently~\cite{mtmsn,Segal2019ASA}. For Boolean questions, the challenge is identifying good sub-questions as there is a large space of questions such as ``Did musicians work for X?'' that may have the expected yes/no answer but are not part of the true decomposition. Semantic parsing faces similar issues when questions have a large number of possible logical forms~\cite{dasigi-etal-2019-iterative}. Finally, end-to-end training of the \decomposer and QA models via REINFORCE~\cite{reinforce} can further improve the score and allow for faster greedy inference.

\subsection*{Acknowledgements}
We thank the Aristo team at AI2 for helpful input, Beaker team for their support with experiments, Dirk Groeneveld for providing the output of the Quark system for evaluation, and Jonathan Berant, Matt Gardner, and Hanna Hajishirzi for invaluable feedback on initial drafts of this paper.

\bibliographystyle{acl_natbib}
\bibliography{modularqa}

\clearpage
\appendix

\begin{table*}[htbp]
    \centering
    \small
    \setlength{\tabcolsep}{6pt}
    \begin{tabular}{llccccccc}
        &  & \multicolumn{4}{c}{\drop} & \multicolumn{3}{c}{\hotpot} \\ 
        \cmidrule(lr){3-6} \cmidrule(lr){7-9} 
        & & All & Diff & Comp & Cmpl & All & Bridge & Comp. \\
        & \textbf{E\;\;\;G\;\;\;R} & EM | F1 & F1 & F1 & F1 & EM | F1 & EM | F1 & EM | F1 \\
        \midrule
     \modqarow \name\; &\tick\; \tick\; \tick   &  86.6 | 87.9 & 85.2 & 81.0 & 96.6 & 48.5 | 61.8 & 50.7 | 64.9 & 39.8 | 49.2 \\
    BART  & \notick\; \tick\; \notick   &  26.7 | 28.0 & \phantom{0}7.7 & 77.5 & 12.8 & 38.0 | 49.0 & 40.4 | 52.9 & 28.0 | 33.7 \\ 
    \midrule
    NMN-D$\dagger$\; & \tick\; \notick\; \notick   &  71.0 | 79.1* & -- & -- & -- & -- & -- & --  \\
        NumNet+V2 & \notick\; \notick\; \notick   &  90.6 | 91.6 & 86.5 & 94.5 & 95.5 & -- & -- & -- \\
        SNMN$\dagger$\; &\tick\; \notick\; \notick   & -- & -- & -- & -- & 50.0 | 63.1 & 48.8 | 63.8 & 54.8 | 60.1 \\ 
        Quark &\notick\; \notick\; \notick  & -- & -- & -- & -- & 61.7 | 75.5 & 62.5 | 78.1 &  58.3 | 64.9 \\
        \bottomrule
    \end{tabular}
    \caption{
    Expanded version of the quantitative part of Table~\ref{tab:results}, reporting both F1 and EM scores in each case. The first three columns, as before, denote qualitative capabilities of each model: whether it can \textbf{E}xplain its reasoning, \textbf{G}eneralize well to multiple datasets, or \textbf{R}e-use existing QA models.
    }
    \label{app:tab:results}
\end{table*}

\section{Model Settings}
\label{app:hyper}
Each \bart-Large model (406M parameters) is trained with the same set of hyper-parameters -- batch size of 64, learning rate of 5e-6, triangular learning rate scheduler with a warmup of 500 steps, and training over 5 epochs. Each \roberta model is trained with the same set of hyper-parameters but a smaller batch size of 16. We selected these parameters based on early experiments and did not perform any hyper-parameter tuning thereafter. All the baseline models are trained with their default hyper-parameters provided by the authors. All the experiments are performed on single-GPU machines (either V100 GPUs or RTX 8000s). 

We always used nucleus sampling to sample sequences from the \bart models. To sample the sub-question using the \squad sub-question generator, we sampled 5 questions for each step  with p=0.95 and max question length of 40. To sample the question decompositions during inference, we additionally set k=10 to reduce the noise in these questions. 

\subsection{Training \squad Question Generator}
We use the \squad 2.0 answerable questions to generate the training data for our \squad question generator. We use the nouns, verbs, nouns, adjectives and adverbs (pos tags=[NOUN, VERB, NUM, PROPN, ADJ, RB]) from the question to define the vocabulary hints (after filtering stop words). To simulate the noisy vocabulary, we also add distractor terms with similar pos tags from other questions from the same paragraph. We sample $j \in [2, ..., 7]$ distractor terms  for each question and add them to the vocabulary hints. 

\subsection{Generating sub-questions}
\label{app:subq}
For every step in the reasoning process, we generate 5 questions using nucleus sampling. We select the questions that the corresponding sub-model is able to answer correctly. For each sub-question, we generate 5 questions in the next step (and so on). At the end, we select all the successful question chains (i.e each sub-question was answered by the sub-model to produce the expected answer at each step).

\subsection{Selecting Question Decompositions}
\label{app:decomp}
It is possible that some of these sub-questions, while valid answerable questions, introduce other words mentioned in the paragraph. However, these may not be valid decompositions of the original question. E.g., for the complex question: "When was the 44th US President born?", the sub-question may state "Who was the 44th President from Hawaii?". While this a valid question with the expected intermediate answer, it introduces irrelevant words that would not be possible for the \decomposer to learn. 

To filter out such potentially noisy decompositions, we compute three statistics based on non-stopword overlap. We compute the proportion of new words introduced in a decomposition $\chain=\{..., \ques_i, \ans_i, ...\ans_n\}$  that were not in the input question or any of the previous answers, that is:
\[
\theta(\chain) = \frac{ \big| \bigcup_i\{w \in \ques_i \mid w \not \in \compq\; \textrm{and} \; \forall j < i \; w \not \in \ans_j\}\big|}{\big| \{w \in \compq\}\big|}
\]
We also compute the number of words from the input question not covered by the decomposition:
\[
\mu(\chain) = \frac{ \big| \{w \in \compq \mid \forall i\;  w \not \in \ques_i \}\big|}{\big| \{w \in \compq\}\big|}
\]

Lastly, we compute the number of answers $\nu$ that were not used in any subsequent question, i.e., the sub-question associated with this answer is irrelevant:
\begin{align*}
\nu(\chain) = \big|\{\ans_i \mid \neg (\exists w \in \ans_i\;
 \textrm{s.t.}\; & w \in \ques_j\; \textrm{where}\; j > i\;\\ & \textrm{or}\; w \in \ans_n) \}\big|
\end{align*}

We only select the  decompositions where $\theta < 0.3$, $\mu < 0.3$, $\theta + \mu < 0.4$, and $\nu=0$. To prevent a single question from dominating the training data, we select upto 50 decompositions for any input question. These hyper-parameters were selected early in the development and gave reasonable results. Minor variations did not have a substantial impact and hence were not tuned on the target set.

\begin{figure*}
    \centering
    \includegraphics[width=\linewidth,trim=0cm 0.3cm 0cm 0.5cm]{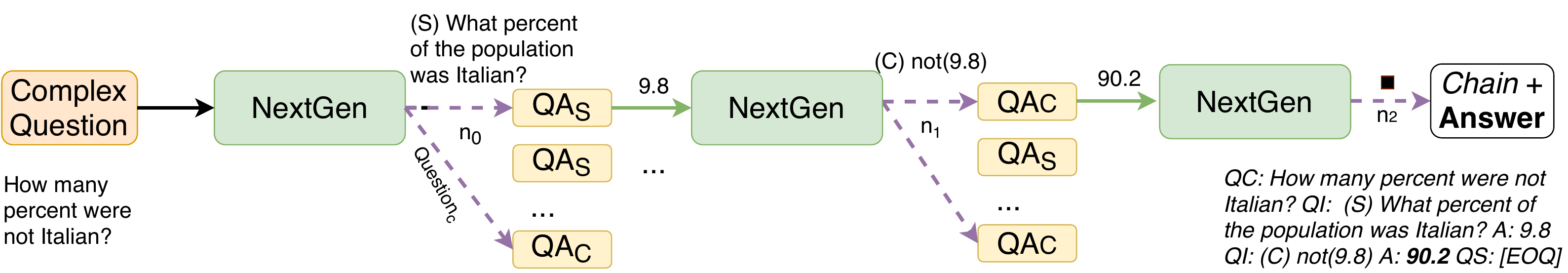}
    \caption{A sample inference chain scored by \name\ for a negation \drop question. For each $n_0$ question generated in the first step, we will explore $n_1$ questions in the second step (and so on). We use our scoring function $w$ to select the optimal inference chain+answer (and prune incomplete low-scoring chains).}
    \label{fig:inf_flow}
\end{figure*}

\subsection{Inference Parameters}
\label{app:inference}
We sample $n_i$ questions in the $i^{th}$ question decomposition step. To ensure sufficient exploration of the search space, we initially sample a larger number of questions but scale them down every step for efficiency. Due to the pipeline nature of our system, it is difficult for our model to recover from any missed question early in the search. We set the number of sampled questions as $n_i = N*r^{i}$ where N=15 and $r=\frac{1}{2}$. When we use greedy inference, we just sample one most-likely question using beam search with width=4. For the QA models, we always select the most likely answer. When there are multiple input paragraphs (e.g., \hotpot), we run the QA model against each paragraph independently and select the most likely answer based on the probability.

To score each generated question, we again rely on the same word-overlap statistic used to filter decompositions. We only use the $\theta$ metric that captures the number of new words introduced in a question chain. The other two metrics are non-motonic i.e they could go down depending on future questions and answers in the chain. At the end, we use a chain scorer (described next) to score each decomposition chain. While we use the $\theta$ metric to guide the search, we primarily rely on the chain score $\delta$ to select the right answer. As a result, the final score for a chain $\chain$ is a weighted combination of these two metrics with higher emphasis on $\delta$
\[
\textrm{score}(\chain) = \theta(\chain) + \lambda \delta(\chain)
\]
where $\lambda$=10 (was set initially during development and not fine-tuned). $\delta$ can only be computed for a complete decomposition and is set to zero for the intermediate steps. Note that higher this score, the worse the chain i.e. we need to find the chain with the lowest score. This scoring function is monotonically increasing as any continuation of a chain will have the same or higher score. We can thus ignore any partial chains with higher scores once we find a complete chain with the lowest score.

\begin{table*}[ht]
\centering
\small
\begin{tabular}{p{2.2cm}p{5cm}p{8cm}}
Operation & QA Model, $\qamodel_\mathqabb$ & Question Generator, $\qgenmodel_\mathqabb$\\\toprule
   \diffc(X, Y, [Z]) & Return absolute difference between X and Y. If $Z \in$ \{days, months, years\}, find the difference in Z. & Generate questions with all possible date/number pairs as X,Y. If $Z \in$ \{days, months, years\} is mentioned in the question, add Z \\
   \notc(X) & Return 100 - X & Generate questions for every number $\leq$ 100 as X.\\
     \ifthenc(X [<>$\neq$] Y, Z, W) & If X is [<>$\neq$] Y, return Z else return W & Generate questions with all possible date/number pairs as X and Y. Use pair of entities in the question as Z and W. \\
    \bottomrule
\end{tabular}
\caption{Set of operations handled by the symbolic calculator model $\qamodel_\mathqabb$ and the corresponding approach to generate such questions in $\qgenmodel_\mathqabb$.}
\label{tab:mathqa}
\end{table*}


\subsection{Chain Scorer}
\label{app:chain_scorer}
To train the scorer, we first collect positive and negative chains by running inference with just the $\theta$ metric. For efficiency, we set the inference parameter N=5 here. For every complete chain, we compute the F1 score of the final answer with the gold answer. If the F1 score exceeds a threshold (0.2 in our case), we assume this chain to be a positive example. We collect such positive and negative chain examples from the training set and then train a \roberta model to classify these chains. We use the \roberta model's predicted probability for the negative class as the score $\delta$.

\subsection{Less Data Training}
\label{app:less_data}
Since we sample the training data for our \decomposer, we can generate more training data by sampling more questions. When training on 20\% of the training data, i.e. only 2600 questions, we sample 15 questions at each step when we are generating the sub-questions(App.~\ref{app:subq}). Similarly we increase N=10 to generate more chains for the chain scorer(App.~\ref{app:chain_scorer}). 

\section{Additional Results}
\label{app:results}

Table~\ref{app:tab:results} expands upon the quantitative results in Table~\ref{tab:results} and reports both F1 and EM (exact match) scores in each setting considered.

\section{Human Evaluation}
\label{app:human_eval}
To identify a subset of questions that \emph{might} be within the scope of our system, we used the human-authored BREAK decompositions~\cite{break} and filtered out questions that require Boolean operations or list operations. We identified the former by the presence of patterns such as "Which is true" and latter by the presence of plural return terms in the sub-questions.

On this resulting subset of 253 questions, \name is comparable to the DecompRC system with only a 2pt F1 gap. Out of the 253 in-scope dev questions, DecompRC did not produce a chain at all (i.e., relied on single-hop reasoning) for 79 questions, whereas \name\ did produce a chain. In contrast, \name\ failed on only 12 questions on which DecompRC succeeded in producing a chain (both systems failed on 3 questions). 

We used crowdworkers to annotate the generated explanations\footnote{We normalized explanations from both the systems for a fair comparison, e.g., lower-cased our explanations, used the system's answers for both, and converted symbolic terms in both explanations to natural language.} on 155 questions where both systems produced a chain. The annotation covered three dimensions:
(1) given the explanation, which system's answer do they \textbf{trust}; (2) which system's explanation do they \textbf{understand} better; and (3) overall, which system's explanation do they \textbf{prefer} (subjective).

\section{Hints for Complex QA Tasks}
\label{app:intera}

\begin{table*}[htbp]
    \centering
    \small
    \begin{tabular}{p{3.5cm}p{3.5cm}p{8cm}}
        Complex Q & Input Hints \hint & Output Sub-Questions \\
        \toprule
         \multirow{3}{*}{\parbox[t]{3.5cm}{\compq: \textit{How many years did it take the services sector to rebound after the 2002 decrease?}}} & \parbox[t]{3.8cm}{$\ans_1$: \anse{2002}, $\vhint_1$:  \vhintf(\compq), $\context_1$: \context} &  
        \parbox[t]{8cm}{$\ques_1$: \quese{When did the services sector take a decrease?}; 
          $\submodel_1$: \squadabb}\\
          \cmidrule{2-3}
          & \parbox[t]{3.8cm}{$\ans_2$: \anse{2003}, $\vhint_2$:  \vhintf(\compq), $\context_2$: \context} &  
        \parbox[t]{8cm}{$\ques_2$: \quese{When did the services sector rebound?};
          $\submodel_2$: \squadabb}\\
          \cmidrule{2-3}
          & \parbox[t]{3.5cm}{$\ans_3$: \anse{1}, $\vhint_3$:  [``diff'', ``2002'', ``2003''], $\context_3$: \context} &  
        \parbox[t]{5cm}{$\ques_3$: \quese{\diffc(2002, 2003)};
          $\submodel_3$: \mathqabb}\\
          \midrule
          \multirow{3}{*}{\parbox[t]{3.5cm}{\compq: \textit{Which ancestral group is smaller: Irish or Italian?}}} & \parbox[t]{3.8cm}{$\ans_1$: \anse{12.2}, $\vhint_1$:  \vhintf(\compq), $\context_1$: \context} &  
        \parbox[t]{8cm}{$\ques_1$: \quese{How many of the group were Irish?}; 
          $\submodel_1$: \squadabb}\\
          \cmidrule{2-3}
          & \parbox[t]{3.8cm}{$\ans_2$: \anse{6.1}, $\vhint_2$:  \vhintf(\compq), $\context_2$: \context} &  
        \parbox[t]{8cm}{$\ques_2$: \quese{How many Italian were there in the group?};
          $\submodel_2$: \squadabb}\\
          \cmidrule{2-3}
          & \parbox[t]{3.5cm}{$\ans_3$: \anse{1}, $\vhint_3$:  [``\ifthenc'', ``12.2'', ``6.1''], $\context_3$: \context} &  
        \parbox[t]{5cm}{$\ques_3$: \quese{\ifthenc(12.2 < 6.1, Irish, Italian)};
          $\submodel_3$: \mathqabb}\\
          \midrule
          \multirow{2}{*}{\parbox[t]{3.5cm}{\compq: \textit{How many percent of the national population does not live in Bangkok?}}} & \parbox[t]{3.5cm}{$\ans_1$: \anse{12.6}, $\vhint_1$:  \vhintf(\compq), $\context_1$: \context} &  
        \parbox[t]{7cm}{$\ques_1$: \quese{What percent of the national population lives in Bangkok?};
          $\submodel_1$: \squadabb}\\
          \cmidrule{2-3}
          & \parbox[t]{3.5cm}{$\ans_2$: \anse{87.4}, $\vhint_2$:[``\notc'', ``12.6'']   , $\context_2$: \context} &  
        \parbox[t]{7cm}{$\ques_1$: \quese{\notc(12.6)};
          $\submodel_2$: \mathqabb}\\
          \midrule
          \multirow{2}{*}{\parbox[t]{3.5cm}{\compq: \textit{Little Big Girl was a Simpsons episode directed by the animator and artist of what nationality?}}} & \parbox[t]{3.5cm}{$\ans_1$: \anse{Raymond S Persi}, $\vhint_1$:  $\zeta(\compq, d_1, d_2)$, $\context_1$: $d_1$} &  
        \parbox[t]{7cm}{$\ques_1$: \quese{Who directed ``Little Big Girl''?};
          $\submodel_1$: \squadabb}\\
          \cmidrule{2-3}
          & \parbox[t]{3.5cm}{$\ans_2$: \anse{American}, $\vhint_2$:   $\zeta(\compq, d_2, d_1)$+ $\ans_1$, $\context_2$: $d_2$} &  
        \parbox[t]{7cm}{$\ques_1$: \quese{What nationality was Raymond S?};
          $\submodel_2$: \squadabb}\\
          
    \end{tabular}
    \caption{Sample hints and the resulting decomposition for \drop and \hotpot examples. The function $\vhintf$ selects non-stopword words and $\zeta(q, d_1, d_2)$ selects the words from $\vhintf(q)$ that don't exclusively appear in $d_2$. 
    }
    \label{tab:decomp_eg}
\end{table*}


To apply \generalname\ to any complex QA dataset, we need to be able to extract the hints needed by the \qgen. As mentioned earlier, these need not have full coverage or have 100\% precision. 

\subsection{HotpotQA}
The questions \compq\ in \hotpot have two supporting gold documents: $d_1$ and $d_2$. 
 Additionally they are also partitioned into two classes: Bridge and Comparison questions.

\subsubsection{Bridge Questions}
There are two forms of bridge questions in \hotpot:
\paragraph{Composition questions:} 
These questions need to first find an intermediate entity $e_1$ that is referred by a sub-question in \hotpot. This intermediate entity points to the final answer through the second sub-question. Generally this intermediate entity is the title entity of the document containing the answer. Say $d_2$ is the document containing the answer and $d_1$ is the other document. If we are able to find a span that matches the title of $d_2$ in $d_1$ and the answer only appears in $d_2$, we assume it to be a composition question. We set $e_1$ to the span that matches the title of $d_2$ in $d_1$.

For the question vocabulary, we could use the terms from the entire question for both steps. Also the second sub-question will use the answer of the first sub-question, so we add it to the vocabulary too.  However, we can reduce some noise by removing the terms that are exclusively appear in the other document. The final hints for this question are:
\begin{align*}
  \context_1 & = d_1; \ans_1=e_1; \hint_1=\zeta(\compq, d_1, d_2) \\
  \context_2 & = d_2; \ans_2=\ans; \hint_2=\zeta(\compq, d_2, d_1) + e_1
\end{align*}
where $\zeta(\ques, d_1, d_2)$ indicates the terms in \ques\ that appear in $d_2$ but not in $d_1$.

\paragraph{Conjunction questions:}
These class of questions do not have any intermediate entity but have two sub-questions with the same answer e.g. ``Who is a politician and an actor?''. If the answer appears in both supporting paragraphs, we assume that it is a conjunction question. The hints for such questions are simple:
\begin{align*}
  \context_1 = d_1; \ans_1=\ans; \hint_1=\zeta(\compq, d_1, d_2) \\
  \context_2 = d_2; \ans_2=\ans; \hint_2=\zeta(\compq, d_2, d_1)
\end{align*}

\subsection{Comparison Questions}
These questions compare certain attribute between two entities/events mentioned in the question. E.g., ``Who is younger: X or Y?''. We identify the two entities $e_1$ and $e_2$ in such questions and find dates/numbers that are mentioned in documents. For every $n_1$, $n_2$ number/date mentioned in the document $d_1$ and $d_2$ respectively, we create the following hints:
\begin{align*}
  & \context_1 = d_1; \ans_1=n_1; \hint_1=\zeta(\compq, d_1, d_2) \\
  & \context_2 = d_2; \ans_2=n_2;  \hint_2=\zeta(\compq, d_2, d_1) \\
  & \context_3 = \phi; \ans_3=\ans; \hint_3=[\ifthenc, n_1, n_2, e_1, e_2]
\end{align*}
The final set of hints would be used by the calculator generator to create the questions: $\ifthenc(n_1 > n_2, e_1, e_2)$ and $\ifthenc(n_1 < n_2, e_1, e_2)$.

\subsection{DROP}
For the questions in \drop, we first identify the class of question that it may belong to and then generate the appropriate hints. Note that one question can belong to multiple classes and we would generate multiple sets of hints in such cases. The questions \compq\ in \drop have only one associated context \context.

\subsubsection{Difference Questions}
We identify these questions based on the presence of term indicating a measurement: "how many" and terms indicating difference such as ``shorter', ``more'', ``days between'', etc. We remove questions that match patterns indicating counting or minimum/maximum such as ``shortest'', ``how many touchdown'', etc. Table~\ref{tab:diff_regexes} shows the regexes that must match and ones that must not match for a question to be categorized as a difference question. 

Finally we check for two dates or numbers in the context such that their difference (in all units) can lead to the final answer. If these conditions are satisfied, for every pair $n_1$, $n_2$ where the difference (in units u) can lead to the final answer, we generate the hints:
\begin{align*}
  & \context_1 = \context; \ans_1=n_1; \vhint_1=\vhintf(\compq) \\
  & \context_2 = \context; \ans_2=n_2;  \vhint_2=\vhintf(\compq) \\
  & \context_3 = \phi; \ans_3=\ans; \vhint_3=[\diffc, n_1, n_2, u]
\end{align*}


\begin{table*}[ht]
    \centering
    \small
    \begin{tabular}{p{8cm}|p{8cm}}
     \textbf{Must match} & \textbf{Should not match} \\
     \hline
     \T \parbox{8cm}{ ".*how many (days|months|years).*",
    ".*how many.*(days|months|years).* between .*",
    ".*how many.* shorter .+ than .*",
    ".*how many.* shorter .+ compar.*",
    ".*how many.* longer .+ than .*",
    ".*how many.* longer .+ compar.*",
    ".*how many.* less .+ than .*",
    ".*how many.* less .+ compar.*",
    ".*how many.* more .+ than .*",
    ".*how many.* more .+ compar.*",
    ".*difference.*"} & 
    \parbox{8cm}{".*minimum.*",
    ".*maximum.*"
    ".*longest.*",
    ".*shortest.*",
    ".*highest.*",
    ".*lowest.*",
    ".*first.*",
    ".*last.*",
    ".*second.*",
    ".*third.*",
    ".*fourth.*",
    ".*how many touchdown.*",
    ".*how many field goal.*",
    ".*how many point.*",
    ".*more touchdown.*",
    ".*more field goal.*",
    ".*more point.*"}
    \end{tabular}
    \caption{Regexes used to identify difference questions and filter out false positives.}
    \label{tab:diff_regexes}
\end{table*}


\begin{table*}[ht]
    \centering
    \small
    \renewcommand{\arraystretch}{1}
    \begin{tabular}{l|l|p{8cm}|r}
    \hline Type & Sub-Type & Description & \#Errs \\
    \hline
    \T    \multirow{6}{*}{Decomposition} &  Low Score & \roberta chain scorer model returned a lower score for the correct decomposition & 5 \\
        & Incomplete  & The decomposition missed a key part of the complex question (e.g. "When did Killgrew marry?" instead of "When did Killgrew marry Catherine?") & 3 \\
        & Sampling  & A correct and better scoring decomposition  exists but was not generated during the search   & 3 \\
        & Long Question & A sub-question exceeded the max token length of 40 set during question generation & 2 \\
        & Missed Decomp & Valid decomposition was not generated & 2 \\
        & Noisy Q & Questions in the generated decomposition were ill-formed & 1 \\
        \hline
        \T \multirow{3}{*}{QA} & Incorrect Ans &  \squad QA model produced an incorrect answer &  3\\
        & No Answer  &  \squad QA model produced no answer i & 2 \\
        & Partial Answer  & \squad QA model produced a partial answer span & 1 \\
        & MathQA format mismatch & Math QA model was unable to handle input format (e.g. \texttt{\ifthenc(1683-99 $>$ 1591-92,...} & 4 \\
        \hline
        \T Out-of-scope & & Question can not be handled by our sub-models & 2 \\
        \hline \T
        Dataset & & Question makes assumptions not stated in text & 2 \\
    \end{tabular}
    \caption{Break down of errors in 30 DROP questions incorrectly answered (F1 < 0.5) by \name}
    \label{tab:drop_errs}
\end{table*}

\subsubsection{Comparison questions}
We identify these questions based on the presence of the pattern: ``ques: $e_1$ or $e_2$''(specifically we match ``\texttt{([\^{},]+)[:,](.*) or (.*)\textbackslash?}''). We handle them in exactly the same way as \hotpot. Since \drop contexts can have more dates and numbers, we select numbers and dates that are close to the entity mentioned~\cite{gupta2020neural}.
\begin{align*}
  & \context_1 = \context; \ans_1=n_1; \vhint_1=\vhintf(ques) + e_1 \\
  & \context_2 = \context; \ans_2=n_2;  \vhint_2=\vhintf(ques) + e_2 \\
  & \context_3 = \phi; \ans_3=\ans; \vhint_3=[\ifthenc, n_1, n_2, e_1, e_2]
\end{align*}

\subsubsection{Complementation questions}
We identify these questions purely based on the presence of ``.* not .*'' in the question(specifically we match ``\texttt{\^{}(.*percent.*)(\textbackslash Wnot\textbackslash W|n't\textbackslash W)(.*)\$}``) and a number in the context $n_1$ such that the $\ans=100-n_1$. The hints are pretty straightforward too:
\begin{align*}
  & \context_1 = \context; \ans_1=n_1; \vhint_1=\vhintf(\compq) \\
  & \context_2 = \phi; \ans_2=\ans; \vhint_2=[\notc, n_1]
 \end{align*}

\section{Drop Error Analysis}
\label{app:drop_err}
See Table~\ref{tab:drop_errs} for the different error types and their counts. Since the search does not explore all possible decompositions, it is possible that there are other decompositions not considered in this analysis. For example, we marked a question to have an error due to ``sampling'' if running inference again found a higher scoring, valid decomposition that led to the correct answer. However, it is possible that an exhaustive search would find an invalid decomposition with an even lower score. Similarly the error cases due to "Incomplete" decomposition or "No Valid" decomposition could also be due to sampling issues.

\end{document}